\documentclass[letterpaper, 10 pt, conference]{ieeeconf}  

\IEEEoverridecommandlockouts                              

\usepackage{graphicx}
\usepackage{booktabs}
\usepackage{array}
\usepackage{colortbl}
\usepackage{xcolor}
\usepackage{tabularx}
\usepackage{amsmath}
\usepackage{amssymb}
\usepackage{url} 

\definecolor{others}{rgb}{0,0,0} 
\definecolor{barrier}{RGB}{255,120,50}
\definecolor{bicycle}{RGB}{255,192,203}
\definecolor{bus}{rgb}{255, 255, 0}        
\definecolor{car}{rgb}{0, 150, 245}        
\definecolor{construction_vehicle}{RGB}{0, 255, 255}  
\definecolor{motorcycle}{RGB}{200, 180, 0} 
\definecolor{pedestrian}{RGB}{255, 0, 0}   
\definecolor{traffic_cone}{RGB}{255, 240, 150}  
\definecolor{trailer}{RGB}{135, 60, 0}     
\definecolor{truck}{RGB}{160, 32, 240}     
\definecolor{driveable_surface}{RGB}{255, 0, 255} 
\definecolor{other_flat}{RGB}{175, 0, 75}  
\definecolor{sidewalk}{RGB}{75, 0, 75}     
\definecolor{terrain}{RGB}{150, 240, 80}   
\definecolor{manmade}{RGB}{230, 230, 250}  
\definecolor{vegetation}{RGB}{0, 175, 0}   
\definecolor{free}{RGB}{255, 255, 255}     

\title{\LARGE \bf
GSRender: Deduplicated Occupancy Estimation via Weakly Supervised
3D Gaussian Splatting
}

\author{Qianpu Sun$^{1,2}$, Changyong Shu$^{2}$, Sifan Zhou$^{2}$, Runxi Cheng$^{1}$, Yongxian Wei$^{1}$, Zichen Yu$^{3}$, Dawei Yang$^{2}$, \\
Sirui Han$^{4*}$, Yuan Chun$^{1*}$%
\thanks{$^{1}$Qianpu Sun, Runxi Cheng, Yongxian Wei and Yuan Chun are with Tsinghua Shenzhen International Graduate School, Tsinghua University, China.}%
\thanks{$^{2}$Qianpu Sun, Changyong Shu, Sifan Zhou, and Dawei Yang are with Houmo AI, China.}%
\thanks{$^{3}$Zichen Yu is with Dalian University of Technology, China.}%
\thanks{$^{4}$Sirui Han is with The Hong Kong University of Science and Technology, Hong Kong SAR, China.}%
\thanks{*Corresponding author: Yuan Chun, Sirui Han.}%
}

\begin{document}

\maketitle
\thispagestyle{empty}
\pagestyle{empty}

\begin{abstract}

Weakly-supervised 3D occupancy perception is crucial for vision-based autonomous driving in outdoor environments. Previous methods based on NeRF often face a challenge in balancing the number of samples used. Too many samples can decrease efficiency, while too few can compromise accuracy, leading to variations in the mean Intersection over Union (mIoU) by \textbf{5-10 points}. Furthermore, even with surrounding-view image inputs, only a single image is rendered from each viewpoint at any given moment. This limitation leads to duplicated predictions, which significantly impacts the practicality of the approach. However, this issue has largely been overlooked in existing research. To address this, we propose GSRender, which uses 3D Gaussian Splatting for weakly-supervised occupancy estimation, simplifying the sampling process. Additionally, we introduce the \textbf{Ray Compensation} module, which reduces duplicated predictions by compensating for features from adjacent frames. Finally, we redesign the dynamic loss to remove the influence of dynamic objects from adjacent frames. Extensive experiments show that our approach achieves SOTA results in RayIoU (+6.0), while also narrowing the gap with 3D-supervised methods. This work lays a solid foundation for weakly-supervised occupancy perception. The code is available at \url{https://github.com/Jasper-sudo-Sun/GSRender}.

\end{abstract}

\section{INTRODUCTION}

Autonomous driving has advanced rapidly in recent years. As a key task, occupancy estimation has attracted attention from both academia and industry. This task models the scene as a grid-based collection, where each grid is assigned relevant attributes such as semantics and motion flow. Compared to traditional 3D object detection \cite{qian20223d, zhou2023fastpillars, kim2021survey} and BEV(Bird’s Eye View) perception \cite{ma2024vision, he2024vision}, 3D occupancy estimation provides a more fine-grained understanding of the environment and compensates for the inability of BEV space to capture height information. Researchers are committed to advancing the accuracy and real-time efficiency of comprehensive scene estimation, aiming to push the boundaries of understanding and interaction with dynamic environments.

\noindent Most existing approaches \cite{liu2023sparsebev, li2023bevstereo, huang2023tri, huang2022bevdet4d, yu2024panoptic, li2023fb, li2023bevdepth, Tang_2024_CVPR} are heavily dependent on 3D ground truth. As noted in RenderOcc \cite{pan2024renderocc}, generating 30,000 frames of 3D labels requires approximately 4,000 human hours. Moreover, inconsistencies in 3D label generation across benchmarks further complicate real-world applications. Thus, reducing or eliminating reliance on 3D labels is a critical challenge, particularly as the demand for 3D perception in autonomous driving grows.

\begin{figure}[t]
    \includegraphics[width=\linewidth]{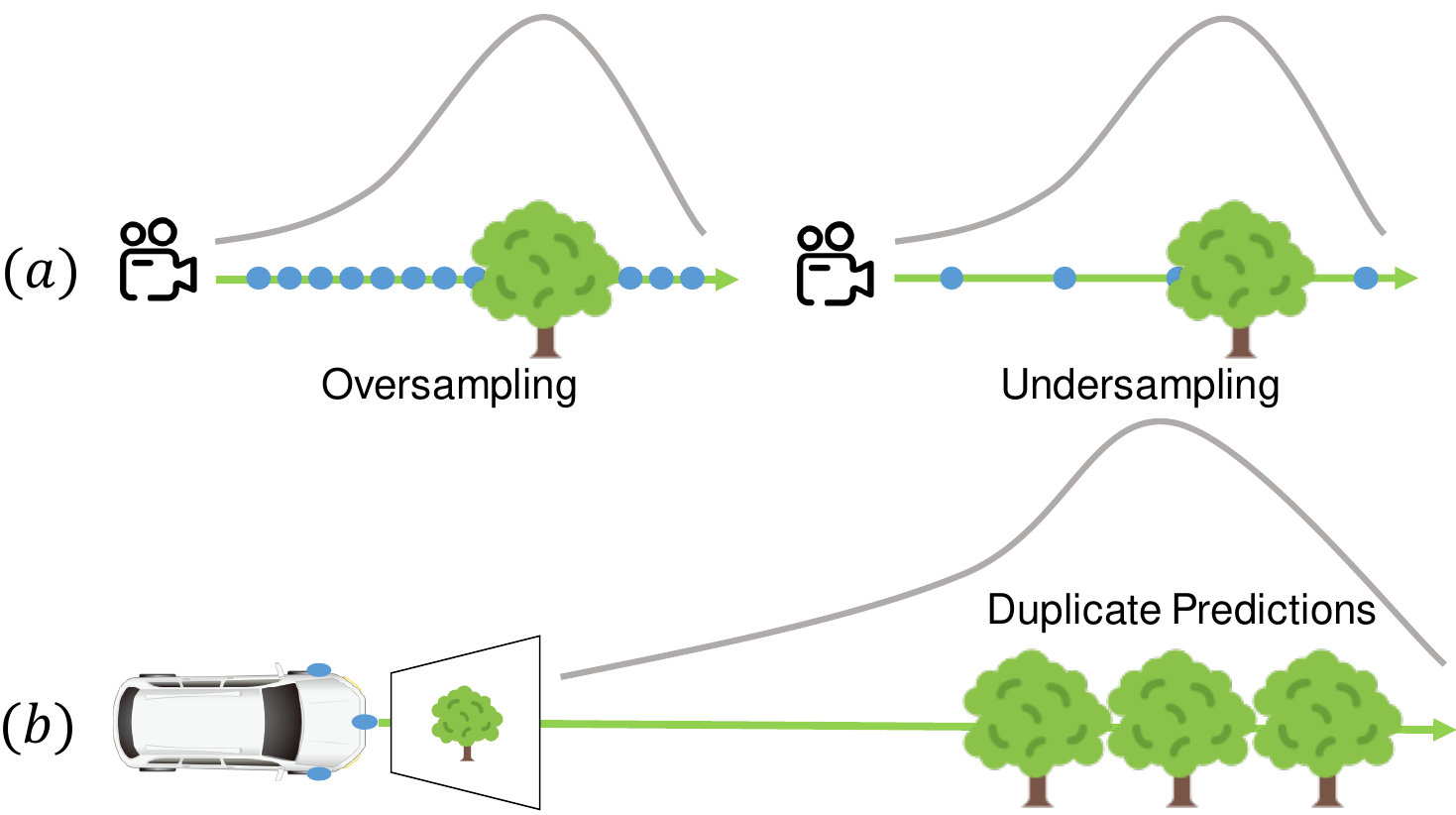}
    \caption{\textbf{The limitations of RenderOcc \cite{pan2024renderocc}}. (a) NeRF-based method face a trade-off between efficiency and precision. (b) Duplicate predictions caused by the uncertainty in depth estimation.}
    \vspace{-0.1cm}
    \label{fig:0}
\end{figure}

\noindent RenderOcc \cite{pan2024renderocc} introduces a method that uses LiDAR point clouds to directly supervise predictions, removing the need for post-processing. Even though RenderOcc is relatively superior, there are still two issues that require our attention, as shown in Figure~\ref{fig:0}. \textbf{(a) Sampling Trade-off:} NeRF-based methods necessitate sampling 3D points along camera rays, but achieving a balanced sampling remains a significant challenge. This complex trade-off may cause accuracy fluctuations ranging from 5 to 10 percentage points. If sampling is insufficient, some positions might be completely undetectable. \textbf{(b) Duplicated Predictions:} Rendering merely relying on single-view images often leads to duplicated predictions along camera rays. Such duplications artificially inflate the mean Intersection over Union (mIoU) metric~\cite{tang2024sparseocc}, thereby reducing the practical utility of 3D occupancy estimation when applied to downstream tasks.

\noindent To address these challenges, we introduce GSRender, an end-to-end framework that leverages 3D Gaussian Splatting \cite{kerbl3Dgaussians}, which enables high-quality rendering of photorealistic scenes by utilizing multiple viewpoint images of the same object. But in occupancy estimation tasks, there are no multi-view images; instead, only surrounding-view images are used, with minimal overlap between different viewpoints. Specifically, we achieve multi-view supervision of the same object by incorporating adjacent frames to address the issue of duplicated predictions. Furthermore, to minimize the potential negative impact of dynamic objects on the rendered scenes, we redesign a differentiated loss for static and dynamic objects, which leads to further performance improvements.

\noindent We conducted all experiments on the nuScenes \cite{caesar2020nuscenes} dataset. It outperforms RenderOcc in the conventional mIoU \cite{long2015fully} metric. We also employed RayIoU \cite{tang2024sparseocc}, a novel metric specifically designed to more accurately measure the model's true predictive performance. Impressively, our method achieved \textbf{SOTA} results in RayIoU under the category of weakly-supervised methods. Specifically, we summarize our contributions as follows:
\begin{enumerate}
    \item We present \textbf{GSRender} and a detailed explanation of the effectiveness of 3D Gaussian Splatting in the analysis section for the weakly-supervised occupancy estimation task.
    \item We propose the \textbf{Ray Compensation module} and a refined loss to mitigate duplicated predictions and handle dynamic and static objects in auxiliary frames, enhancing the practicality of weakly supervised methods.
    \item We achieved \textbf{SOTA} performance in RayIoU metrics among all 2D weakly-supervised methods and substantially reduced the performance gap compared to methods utilizing 3D supervision.
\end{enumerate}

\section{Related Work}

\subsubsection{Voxel-based Scene Representation.} Efficient representations have become a key focus. Voxel-based representations enable fine-grained representations, driving success in LiDAR segmentation \cite{liong2020amvnet, tang2020searching, cheng20212, Tang_2024_CVPR, ye2021drinet++} and 3D scene completion \cite{cao2022monoscene, roldao2020lmscnet, chen20203d}. Since Tesla AI Day 2022 \cite{tesla2022}, occupancy perception has gained increasing attention. Mainstream models fall into two paradigms: the first uses the LSS method \cite{philion2020lift, li2023bevstereo, yu2024panoptic, li2023bevdepth, huang2021bevdet}, which predicts depth within the ego-vehicle's coordinate system, while the second employs Transformer-based methods \cite{vaswani2017attention} \cite{li2023voxformer, li2022bevformer, zhang2023occformer, wei2023surroundocc} with deformable attention \cite{zhu2020deformable} to refine BEV features. However, these methods often rely on dense features, whereas most voxels in the space are empty, leading to computational redundancy in real-world scenarios. Recent works have focused on simplifying the feature space, achieving notable results \cite{tang2024sparseocc, shi2024occupancy, huang2024gaussian, wang2024opus, Tang_2024_CVPR}.

\subsubsection{Gaussian Scene Representation.} 3D Gaussian Splatting \cite{kerbl3Dgaussians} has been applied widely \cite{szymanowicz2024splatter, charatan2024pixelsplat, xu2024grm, yan2024street, shen2024gamba, gao2024gaussianflow, lu20243d, wu20244d}, offering real-time scene reconstruction. It outperforms NeRF in both performance and rendering quality. Recent work treats large-scale outdoor scenes as Gaussian collections, including GaussianFormer \cite{huang2024gaussian}, which enhances each Gaussian using a transformer, and GaussianBEV \cite{chabot2024gaussianbev}, which replaces LSS with Gaussian Splatting for BEV feature extraction. GaussianOcc \cite{gan2024gaussianocc} integrates cross-view information to minimize sensitivity to camera poses. Our research also models outdoor scenes as Gaussians for detailed reconstruction.

\subsubsection{Weakly-Supervised Estimation.} Due to challenges in obtaining 3D labels, many works focus on weakly-supervised paradigms for semantic scene completion. \cite{Wimbauer_2023_CVPR} achieves semantic completion from a single image, and S4C \cite{hayler2024s4c} improves completion by aligning objects across multiple frames. For occupancy estimation, several methods aim to reduce reliance on 3D labels. RenderOcc \cite{pan2024renderocc} uses a NeRF-based approach to bridge semantic and density fields with depth maps from LiDAR. OccNeRF \cite{zhang2023occnerf} and SelfOcc \cite{huang2024selfocc} use self-supervision for generating 2D semantic and depth maps. However, the issue of duplicated predictions from a single viewpoint persists due to the limited overlap between different viewpoints at the same time. We address this issue by introducing temporal supervision, which helps mitigate the problem of duplicated predictions from a single viewpoint.

\begin{figure*}[t!]
    \includegraphics[width=\linewidth]{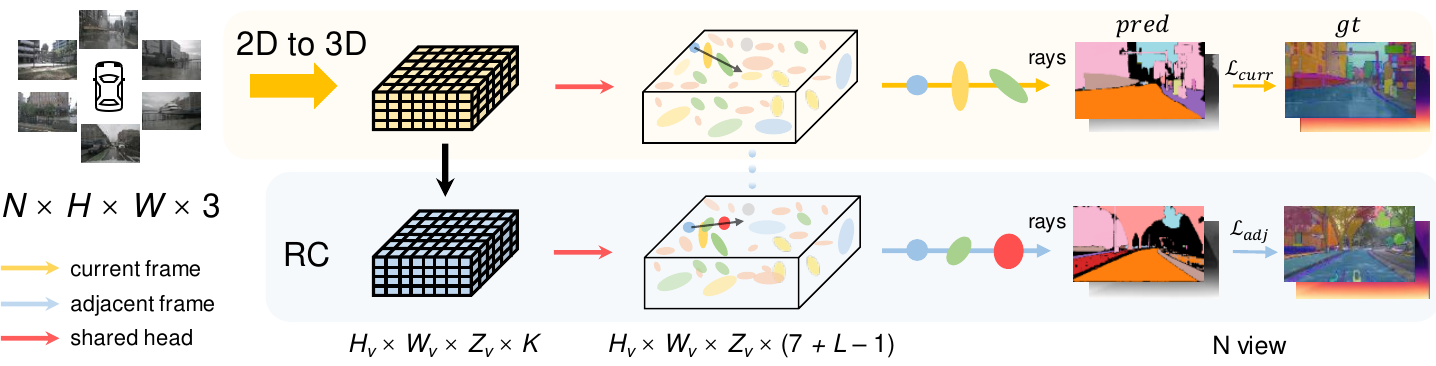}
    \caption{\textbf{Overall Framework of GSRender}. For surround view image input, we employ an arbitrary 2D to 3D module to extract occupancy features. Using a simple Gaussian head, we predict the attributes of each Gaussian, followed by Gaussian rendering. Then, we achieve compensation for different viewpoints of the same object through the Ray Compensation (RC) module, alleviating the issue of duplicate predictions.}
    \label{fig:2}
\end{figure*}

\section{Preliminary}
We first briefly introduce the essential knowledge of 3D Gaussian Splatting, the problem definition, and the evaluation metrics we use.

\subsubsection{3D Gaussian Splatting.} 3D Gaussian Splatting \cite{kerbl3Dgaussians} is a highly popular method in the field of 3D reconstruction, used for real-time reconstruction of both single objects and large-scale scenes. It conceptualizes the entire scene as a collection of Gaussians. Each Gaussian encompasses attributes \textbf{}such as the mean and covariance of the Gaussian distribution, along with opacity and color. The Gaussian probability for each spatial point is calculated as follows:
\begin{equation}
G(x) = e^{-\frac{1}{2}(X-\mu)^T\Sigma^{-1}(X-\mu)},
\end{equation}
where \(\mu\) represents the position of each Gaussian, while the covariance matrix \(\Sigma\) to determine the size and orientation of the Gaussian distribution. Specifically, the covariance matrix can be computed using rotation \(R\) and scale \(S\) matrices by:
\begin{equation}
    \Sigma = RSS^TR^T.
\end{equation}
For independent optimization, we represent the scale as a 3D vector $\textit{s} \in \mathbb{R}^3$ and the rotation as a quaternion $\textit{q} \in \mathbb{R}^4$. Then, the 3D Gaussians are projected onto the 2D imaging plane using transformation matrices $W$ and Jacobian matrices $J$. The covariance matrix of the 2D Gaussian is computed as follows:
\begin{equation}
    \Sigma' = JW \Sigma W^TJ^T.
\end{equation}
Finally, standard alpha-blending is utilized to render the rgb colors of all Gaussians onto the image based on their opacity, and the loss is computed against the image. The color of each pixel is calculated as follows:
\begin{equation}
\begin{aligned}
c(x) = \sum_{k=1}^{K} c_k \alpha_k G^{2D}_k(x) \prod_{j=1}^{k-1} \left(1 - \alpha_j G^{2D}_j(x)\right).
\end{aligned}
\end{equation}
where \(\alpha_k\) represents the opacity of the \(k\)th Gaussian, \(G_{k}^{2D}(x)\) is the Gaussian probability at pixel position \(x\), and \(c_k\) denotes the color of the \(k\)-th Gaussian, represented by spherical harmonic coefficients \cite{fridovich2022plenoxels}. \(K\) denotes the number of Gaussians contributing to pixel \(x\).

\subsubsection{Problem Definition.} we aim to construct a comprehensive 3D occupancy representation of the scene using images captured from surrounding viewpoints. Specifically, for the vehicle at timestamp $t$, we take $N$ surrounding images \( I = \{I_1, I_2, \ldots, I_N\} \) as input and predict the 3D occupancy grid \( V \in \mathbb{R}^{H_V \times W_V \times Z_V \times L} \), where \( H_V \), \( W_V \), and \( Z_V \) denote the resolution of the voxel space, and \( L \) represents the number of categories, including free category for unoccupied space. The formulation for this 3D prediction task is articulated as follows:
\begin{equation}
P = \mathbb{H}(I_1, I_2, ..., I_N),V = \mathbb{F}(P),
\end{equation} 
Specifically, \(\mathbb{H}\) generates an occupancy feature \(P \in \mathbb{R}^{H_V \times W_V \times Z_V \times C}\), where $C$ denote the feature dimension, capturing spatial structure and temporal relationships within the scene.
Following this, the function $\mathbb{F}$ serves as a head that takes the scene feature representation $P$ as input and outputs the final prediction $V$. RenderOcc \cite{pan2024renderocc} employs a NeRF-based rendering head and achieves multi-view alignment through Auxiliary Rays. In GSRender, we replace the head with a Gaussian-based head and refine the auxiliary rays module, implementing a more efficient approach.

\begin{figure}[t!]
    \includegraphics[width=\linewidth]{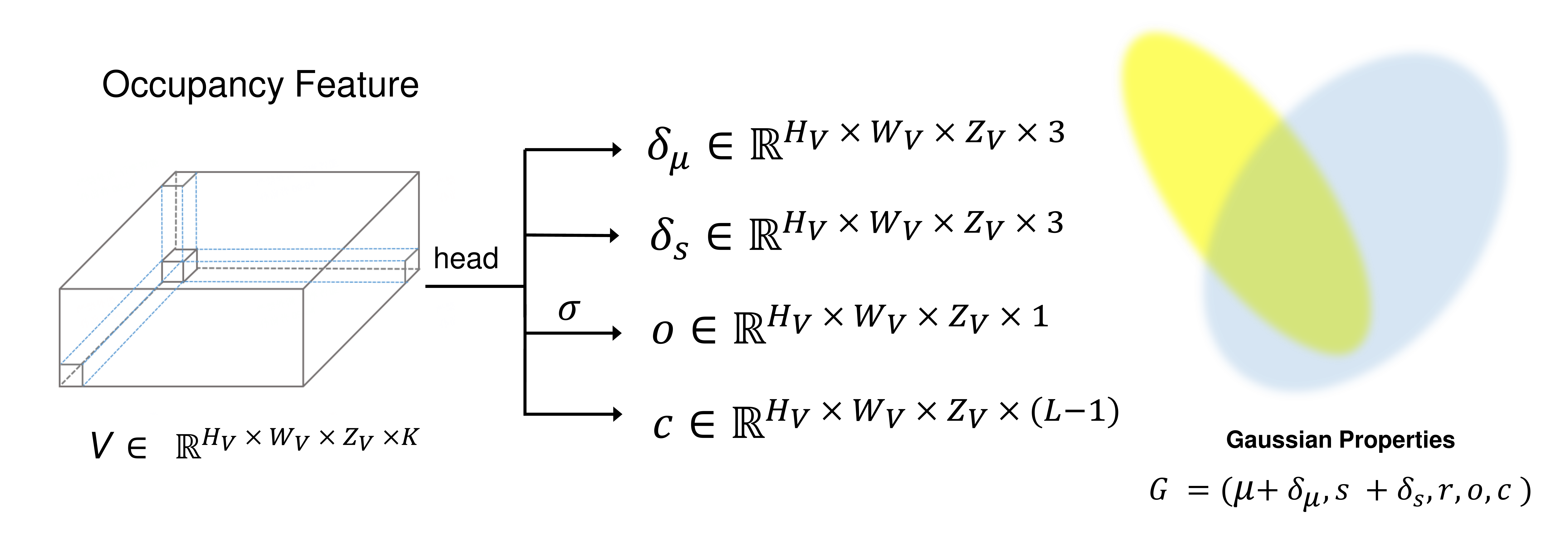}
    \caption{\textbf{Gaussian Properties Field}. The occupancy feature from the 2D to 3D module is fed into the Gaussian head, which outputs the shift of Gaussian's mean \(\delta_\mu\), scales \(\delta_s \), opacity \(o\), and semantic logits \(c\), representing the Gaussian's location, scale, visibility, and semantic category.}
    \vspace{-0.5cm}
    \label{fig:3}
\end{figure}

\
\subsubsection{Evaluation metrics.} We utilized both the mIoU metric and the newly proposed RayIoU \cite{tang2024sparseocc} metric to evaluate our method. The RayIoU metric simulates LiDAR by projecting query rays into the predicted 3D occupancy. A query ray is considered a true positive (TP) if its class matches and the $L1$ depth error between the ground-truth and predicted depths is below a threshold (e.g., 2m). The query ray can originate from the LiDAR position at the current, past, or future moments along the ego path. Temporal casting allows us to evaluate scene completion performance while maintaining a well-posed task. Given C classes, RayIoU is computed as follows:
\begin{equation}
  \text{RayIoU} = \frac{1}{C}\displaystyle \sum_{c=1}^{C}\frac{\text{TP}_c}{\text{TP}_c + \text{FP}_c + \text{FN}_c},
\end{equation}
where $\text{TP}_c$, $\text{FP}_c$, and $\text{FN}_c$ correspond to the number of true positive, false positive, and false negative predictions for class $c_i$. We believe this evaluation method better reflects the model's true performance, so we focus on using it as our primary evaluation metric.

\section{Method}

\subsection{Overall Framework.} 

\noindent Figure~\ref{fig:2} illustrates our overall framework. We conceptualize the entire scene as an assembly of numerous Gaussian. In the feature extraction stage, we adopt the same backbone to ensure a fair comparison with RenderOcc. First, we treat each voxel center as the center of a Gaussian and predict the offset of their properties within the Gaussian. Next, we render the semantics logits and depth and optimize the network with LiDAR rendering supervision. Finally, we introduce a multi-frame calibration module aimed at reducing false positive detections caused by duplicate predictions.

\subsection{Gaussian Properties Field.} We treat each voxel center as the initial point for a gaussian distribution, as shown in Figure~\ref{fig:3}. For a given occupancy feature \(V\), in addition to predicting the semantic and opacity attributes, we initialize the Gaussian's mean and scale, and predict offsets for each property, excluding rotation. In structured voxel spaces, we contend that rotating the Gaussian is unnecessary. Instead, adjusting scale suffices to meet the spatial requirements of the entire voxel space. The regular arrangement of voxels in a uniformly distributed and structured environment already provides sufficient geometric and spatial information, rendering rotational adjustments redundant. Our ablation studies have confirmed this approach. The resulting Gaussian properties \(G \in \mathbb{R}^{N \times (7 + (L-1))}\), where $N$ denotes the total number of Gaussians. For each Gaussian, we have
\begin{equation}
G(i) = (\mu + \delta_{\mu}, s + \delta_s, r, o, c).
\end{equation}
During the testing phase, we treat Gaussians with opacity greater than a predefined threshold as the final predictions. Specifically
\begin{equation}
V(\mu) = 
\begin{cases}
\arg\max \left( C(\mu) \right) & \text{if } o(\mu) \geq \tau \\
\varnothing & \text{if } o(\mu) < \tau
\end{cases}
\end{equation}
where \(\tau\) is the opacity threshold for determining whether a Gaussian is considered empty. Since empty voxels are filtered out based on threshold, the actual number of predicted semantic categories is \(L-1\), excluding the 'free' category. Visualization analysis shows that the majority of effective Gaussians closely align with their voxel centers. Thus, we apply the Gaussian semantics directly to the voxels.

\subsubsection{Gaussian Rasterizer.} We interpret the semantic logits associated with each Gaussian as colors, and apply alpha-blending to render the semantic logits and depth for each pixel. This approach culminates in the generation of a semantic map \( S \) and a depth map \( D \).

\begin{figure}[t!]
    \includegraphics[width=\linewidth]{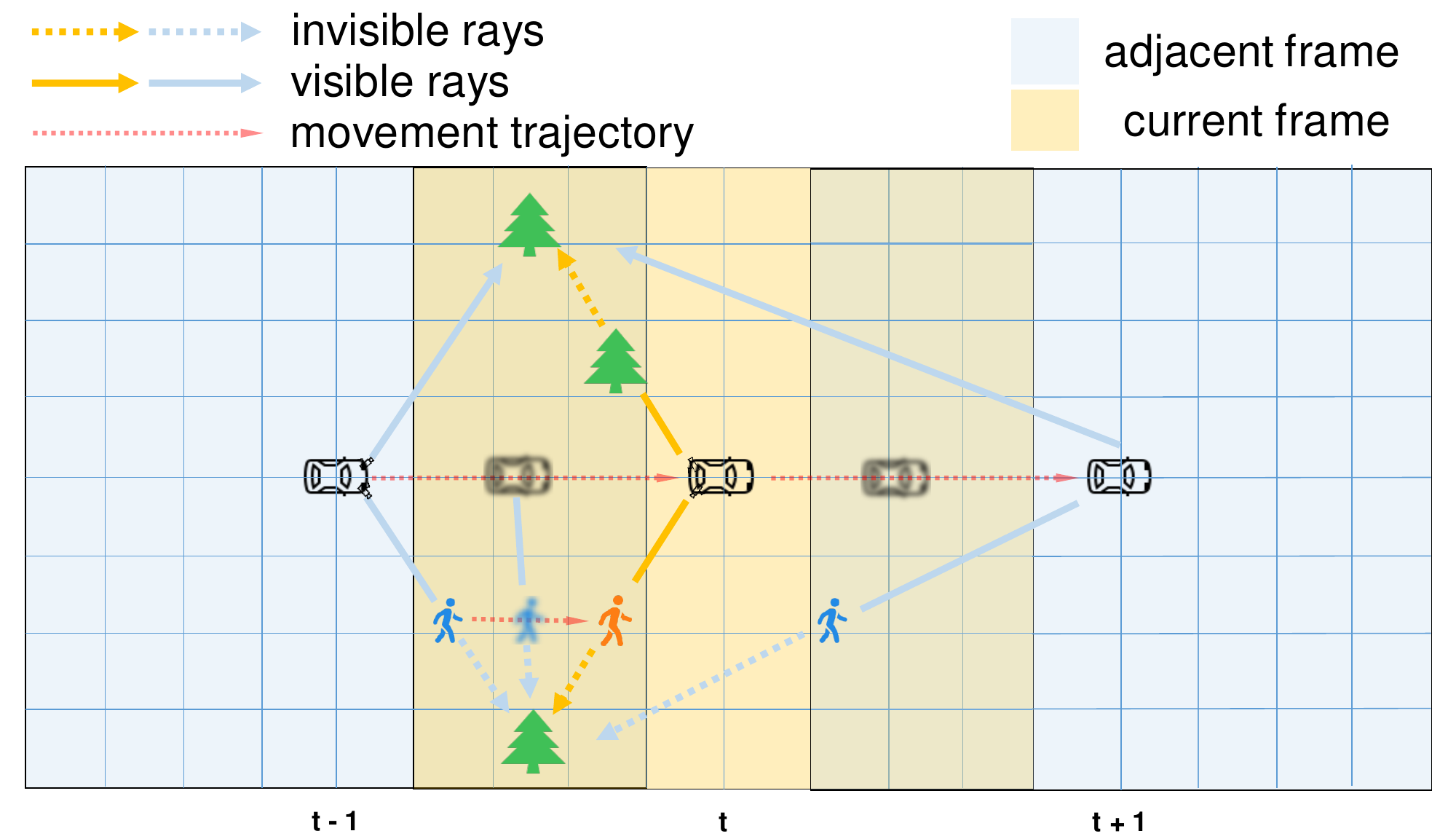}
    \caption{\textbf{Ray Compensation}.  In the upper part of the feature map, it indicates that adjacent frame compensation is used to address occlusion in the current frame. In the lower part of the feature map, it shows that dynamic objects may occlude the view of the compensated frame again, so it's necessary to reduce the contribution of dynamic objects in adjacent frame.}
    \vspace{-0.3cm}
    \label{fig:4}
\end{figure}

\subsection{Ray Compensation Module.} 

\noindent RenderOcc \cite{pan2024renderocc} introduces Auxiliary Rays to ensure spatial consistency across different viewpoints by aligning labeled rays from multiple frames to a single frame. Nevertheless, it fails to resolve the duplicated predictions along these rays. We introduce the Ray Compensation module, which provides multiple perspectives for each object at the feature level. Specifically as shown in Figure~\ref{fig:4}. The upper part of figure illustrates the principle of our method. For the current frame \( t \), trees that are obscured might be revealed in adjacent frames from the different viewpoints. 
To obtain features from adjacent time steps, we first assume that the current Occupancy Feature is denoted as \( P_{\text{curr}} \in \mathbb{R}^{H_V \times W_V \times Z_V \times C} \). We treat the centers of voxels as homogeneous coordinates \( O^{\text{curr\_ego}}_{\text{curr}} \in \mathbb{R}^{H_V \times W_V \times Z_V \times 4} \), representing the voxel centers in the ego-car coordinate system at the current time. The position of the adjacent time step in the current time step’s coordinate system is computed as follows:
\begin{equation}
    O_{\text{adj}}^{\text{curr\_ego}} = T_{\text{adj \_ego}}^{\text{curr \_ego}} O_{\text{adj}}^{\text{adj\_ego}},
\end{equation}
where \( T_{\text{adj\_ego}}^{\text{curr\_ego}} \in \mathbb{R}^{4 \times 4} \) is the transformation matrix from the ego-car coordinate system of the adjacent frame to that of the current frame. The features of adjacent frames are mathematically represented by the following formula:
\begin{equation}
\begin{split}
    P_{adj}^{t + k} = \mathbb{S}(P_{curr}^t, O_{\text{adj}}^{\text{curr\_ego}}), \\
    \text{for} \quad k = -1, 1, \ldots, -\frac{T_{m}}{2}, \frac{T_{m}}{2}.
\end{split}
\end{equation}
Here, \(\mathbb{S}\) denotes the sampling function, where we apply bilinear interpolation, and \(T_m\) represents the total number of adjacent frames selected. Finally, the shared gaussian head predicts the Gaussian properties for the adjacent frame.

\subsubsection{Dynamic Object.} It is noteworthy that this approach may still encounter issues, as illustrated in the lower part of Figure \ref{fig:4}. For trees obscured in the current frame, they may still be blocked by other moving objects in adjacent frames. In extreme cases, these obstructions might occur at every time step. Therefore, we need to minimize the influence of dynamic objects in adjacent frames. This is achieved by applying a weighting factor, alpha, when calculating the loss for adjacent frames, as detailed in the following subsection.

\subsection{Loss Function}
To maintain consistency with RenderOcc, we also optimize the Gaussian properties using a class-balanced cross-entropy loss \(\mathcal{L}^{\text{seg}}\) for occupancy prediction and the SILog loss \(\mathcal{L}^{\text{depth}}\) \cite{eigen2014depth} for depth estimation. For the current frame, we compute the loss similarly to RenderOcc as follows: 
\begin{equation}
    \mathcal{L}_{curr} = \mathcal{L}^{seg}_{curr} + \mathcal{L}^{depth}_{curr}.
\end{equation}
For adjacent frames, the inability to estimate the motion trajectories of moving objects can negatively impact network optimization. Therefore, a lower weight is assigned to dynamic objects. The adjacent loss can be formulated as follows:
\begin{equation}
    \mathcal{L}_{adj} = \mathcal{L}^{seg}_{adj} + \mathcal{L}^{depth}_{adj}.
\end{equation}
The segmentation loss \(\mathcal{L}^{seg}_{adj}\) and depth loss \(\mathcal{L}^{depth}_{adj}\) are defined as follows:
\begin{equation}
    \mathcal{L}^{seg}_{adj} = - \sum_{i=1}^{n} \alpha_{s(i)}  \beta_{s(i)} \mathbf{S}_{(i)} \log(\mathbf{\hat{S}}_{pix}(i)),
\end{equation}

\begin{equation}
    \mathcal{L}^{depth}_{adj} = \sqrt{\frac{1}{n} \sum_{i=1}^{n} \alpha_{s(i)}d_i^2 - \frac{1}{n^2} \left( \sum_{i=1}^{n} \alpha_{s(i)}d_i \right)^2},
\end{equation}
where \( d_i = \log \mathbf{D}_i - \log \hat{\mathbf{D}}_i \) denotes the difference between predicted and ground truth depth at pixel \(i\), while \(s(i)\) is the ground truth semantic category. \( \alpha_{s(i)} \) and \( \beta_{s(i)} \) represent the dynamic weight and category balance weight for $s(i)$, respectively. Consequently, our final loss function can be expressed as:
\begin{equation}
\mathcal{L} = \mathcal{L}_{\text{curr}} + \sum_{k} \omega_k \mathcal{L}_{adj, k}
\end{equation}
where \(\omega\) represents the weight assigned to adjacent frames. Additionally, although we also considered using the distortion loss proposed in 2D Gaussian Splatting \cite{huang20242d}, we found it to be less effective and decided not to include it.


\section{Experiments}
We trained and evaluated our method on the nuscenes dataset, leveraging the surround-view camera setup for better feature interaction. Additionally, we introduced RayIoU \cite{tang2024sparseocc} to highlight the effectiveness of multi-frame compensation and conducted ablation studies for deeper insights into GSRender.

\begin{table*}[htbp]
    \centering
    \resizebox{\textwidth}{!}{  
    \begin{tabular}{cc|c|ccccccccccccccccc}
        \toprule
        method                 & \rotatebox{90}{GT} & \rotatebox{90}{mIoU}  & \rotatebox{90}{\textcolor{others}{\rule{2mm}{2mm}} others} & \rotatebox{90}{\textcolor{barrier}{\rule{2mm}{2mm}} barrier} & \rotatebox{90}{\textcolor{bicycle}{\rule{2mm}{2mm}} bicycle} & \rotatebox{90}{\textcolor{bus}{\rule{2mm}{2mm}} bus}   & \rotatebox{90}{\textcolor{car}{\rule{2mm}{2mm}} car}   & \rotatebox{90}{\textcolor{construction_vehicle}{\rule{2mm}{2mm}} cons. veh} & \rotatebox{90}{\textcolor{motorcycle}{\rule{2mm}{2mm}} motorcycle} & \rotatebox{90}{\textcolor{pedestrian}{\rule{2mm}{2mm}} pedestrian} & \rotatebox{90}{\textcolor{traffic_cone}{\rule{2mm}{2mm}} traffic cone} & \rotatebox{90}{\textcolor{trailer}{\rule{2mm}{2mm}} trailer} & \rotatebox{90}{\textcolor{driveable_surface}{\rule{2mm}{2mm}} truck} & \rotatebox{90}{\textcolor{barrier}{\rule{2mm}{2mm}} dri. sur} & \rotatebox{90}{\textcolor{other_flat}{\rule{2mm}{2mm}} other flat} & \rotatebox{90}{\textcolor{sidewalk}{\rule{2mm}{2mm}} sidewalk} & \rotatebox{90}{\textcolor{terrain}{\rule{2mm}{2mm}} terrain} & \rotatebox{90}{\textcolor{manmade}{\rule{2mm}{2mm}} manmade} & \rotatebox{90}{\textcolor{vegetation}{\rule{2mm}{2mm}} vegetation} \\ \hline
        BEVFormer \cite{li2022bevformer}              & 3D & 23.67 & 5.03   & 38.79   & 9.98    & 34.41 & 41.09 & 13.24     & 16.50      & 18.15      & 17.83        & 18.66   & 27.70 & 48.95    & 27.73      & 29.08    & 25.38   & 15.41   & 14.46      \\
        BEVStereo \cite{li2023bevstereo}              & 3D & 24.51 & 5.73   & 38.41   & 7.88    & 38.70 & 41.20 & 17.56     & 17.33      & 14.69      & 10.31        & 16.84   & 29.62 & 54.08    & 28.92      & 32.68    & 26.54   & 18.74   & 17.49      \\
        OccFormer \cite{zhang2023occformer}              & 3D & 21.93 & 5.94   & 30.29   & 12.32   & 34.40 & 39.17 & 14.44     & 16.45      & 17.22      & 9.27         & 13.90   & 26.36 & 50.99    & 30.96      & 34.66    & 22.73   & 6.76    & 6.97       \\
        SelfOcc \cite{huang2024selfocc} & 2D* & 9.30 & - & 0.15 & 0.66 & 5.46 & 12.54 & 0.00 & 0.80 & 2.10 & 0.00  & 0.00 & 8.25 & 55.49 & - &26.30 & 26.54 & 14.22 & 5.60 \\ 
        OccNeRF \cite{zhang2023occnerf}  & 2D* & 9.54 & - & 0.83 & 0.82 & 5.13 & 12.49 & 3.50 & 0.23 & 3.10 & 1.84 & 0.52 & 3.90 & 52.62 & - & 20.81 & 24.75& 18.45 & 13.19 \\
        GaussianOcc \cite{gan2024gaussianocc}  & 2D* & 9.94 & - & 1.79 & 5.82 & 14.58 & 13.55 & 1.30 & 2.82 &  7.95 & 9.76 & 0.56 & 9.61 & 44.59 & - & 20.10 & 17.58 & 8.61 & 10.29     \\
        \midrule
        RenderOcc(1f) \cite{pan2024renderocc}     & LiDAR-2D & 19.33 & 1.82   & 19.54  & 12.6   & 16.08 & 12.85 & \textbf{8.76}   &  12.14     & 9.33      & 12.94        & 17.60  & 12.13 & 56.25    & 28.83      &  \textbf{36.89}    &  38.82  & 15.08  & 16.98      \\
        GSRender(1f) & LiDAR-2D & \textbf{21.36} &   \textbf{2.44}   &    \textbf{20.13 }   &    \textbf{13.56}    & \textbf{ 19.54}      &  \textbf{18.38}     &   8.35     &     \textbf{18.77}    &   \textbf{ 10.67 } &  \textbf{ 14.81}       &     \textbf{16.63 }     &  \textbf{14.92}     & \textbf{61.74}     &  \textbf{29.77}       &    35.95        &   \textbf{40.56}       &    \textbf{19.83}    &  \textbf{ 17.06}    \\ \midrule
        RenderOcc \cite{pan2024renderocc} & LiDAR-2D+3D & 26.11 & 4.84 & 31.72 & 10.72 & 27.67 & 26.45 & 13.87 & 18.20 & \textbf{17.67} & \textbf{17.84} & 21.19 & 23.25 & 63.20 & 36.42 & 46.21 & 44.26 & 19.58 & 20.72 \\
        GSRender & LiDAR-2D+3D & \textbf{29.56} &\textbf{8.42} & \textbf{34.93} & \textbf{15.12} & \textbf{30.71} & \textbf{29.61} & \textbf{16.70} & \textbf{9.48} & 17.61 & 16.58 & \textbf{23.92} & \textbf{27.24} & \textbf{77.94} & \textbf{39.21} & \textbf{51.69} & \textbf{54.61} & \textbf{23.82} & \textbf{24.92}  \\
        \hline
    \end{tabular}
    }
    \caption{Quantitative Comparison on the Occ3D-nuScenes dataset in mIoU metric. 2D$^*$ denotes the semantic and depth maps generated using a semi-supervised method, rather than by projecting LiDAR point clouds onto the imaging plane. The best results obtained using only 2D supervision are highlighted in bold.}  
    \label{tab:1}  
\end{table*}

\begin{table*}[htbp]
    \centering
    \newcolumntype{a}{>{\columncolor{gray!9}}c}
    \resizebox{\textwidth}{!}{  
    \begin{tabular}{c|c|ccc|acccc}
    \toprule[1pt]
    Method          & GT & Backbone & Input Size & Epoch & RayIoU & RayIoU\textsubscript{1m} & RayIoU\textsubscript{2m} & RayIoU\textsubscript{4m} \\ \midrule[1pt]
    FB-Occ (16f) \cite{li2023fb}    & 3D & R50      & 704$\times2$56  & 90    & 33.5   & 26.7   & 34.1   & 39.7   \\
    SparseOcc(8f) \cite{tang2024sparseocc}      & 3D & R50      & 704$\times$256  & 24    & 34.0   & 28.0   & 34.7   & 39.4 \\ 
    Panoptic-FlashOcc (1f) \cite{yu2024panoptic} & 3D & R50      & 704$\times$256  & 24   & 35.2 & 29.4 & 36.0 & 40.1  \\
    SimpleOcc \cite{gan2023simple}       & 3D & R101      & 672$\times$336  & 12    & 22.5  & 17.0   & 22.7   & 27.9 \\ 
    OccNeRF \cite{zhang2023occnerf}         & 2D* & R101      & 640$\times$384  & 24    & 10.5  & 6.9   & 10.3   & 14.3 \\ 
    GaussianOcc \cite{gan2024gaussianocc}  & 2D* & R101 & 640$\times$384 & 12 & 11.9 & 8.7 & 11.9 & 15.0 & \\ 
    \midrule[1pt]
    RenderOcc(2f)   & LiDAR-2D & Swin-B   & 1408$\times$512 & 12    & 19.3   & 12.7   & 19.3   & 25.9   \\ 
    RenderOcc(7f)   & LiDAR-2D & Swin-B   & 1408$\times$512 & 12    & 19.5   & 13.4   & 19.6   & 25.5   \\ 
    GSRender(2f)   & LiDAR-2D & Swin-B & 1408$\times$512 & 12    & \textbf{25.5}\textcolor{green}{(+6.0)}  & \textbf{18.7}\textcolor{green}{(+5.3)}   & \textbf{25.8}\textcolor{green}{(+6.2)}   & \textbf{31.8}\textcolor{green}{(+6.3)}    \\ 
    \bottomrule[1pt]
    \end{tabular}
    }
     \caption{Quantitative Comparison on the Occ3D-nuScenes dataset in RayIoU metric. The best results obtained using only LiDAR-2D supervision are highlighted in bold.}
    \label{tab:2}  
\end{table*}

\subsection{Experimental Setup}
\subsubsection{Dataset.} All of our experiments were conducted on the nuscenes dataset~\cite{caesar2020nuscenes} and the corresponding occupancy dataset, OCC3D-NuScenes~\cite{tian2024occ3d}. The occupancy dataset contains 600 scenes for training and 150 scenes for validation. It covers a region centered around the ego-vehicle with dimensions $[-40\text{m}, -40\text{m}, -1\text{m}, 40\text{m}, 40\text{m}, 5.4\text{m}]$ and uses a voxel size of $[0.4\text{m}, 0.4\text{m}, 0.4\text{m}]$. This dataset contains 18 classes, where classes 0 to 16 follow the same definitions as the nuScenes-lidarseg dataset, and class 17 represents free space. We followed the approach used in RenderOcc \cite{pan2024renderocc} to project the semantic and depth labels of 3D LiDAR points onto the 2D maps.
\begin{figure*}[t!]
    \includegraphics[width=\linewidth]{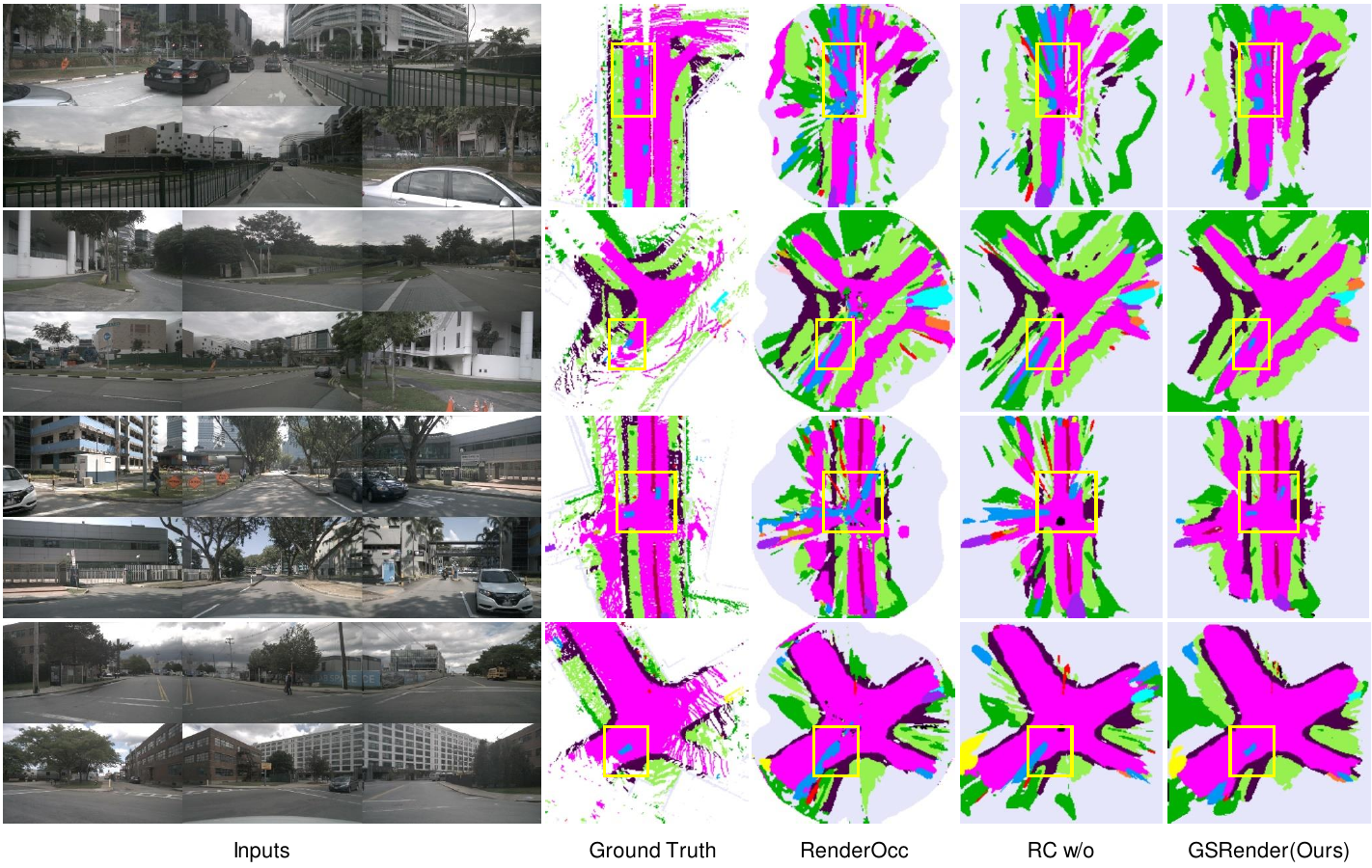}
    \caption{\textbf{Qualitative results of GSRender}. Top view on Occ3D-nuScenes}
    \vspace{-0.2cm}
    \label{fig:5}
\end{figure*}
\subsubsection{Implementation details.} 
We implemented our model using PyTorch. We used off-the-shelf architecture Swin Transformer \cite{liu2021swin} as the image backbone with an input resolution of 512\(\times\)1408 and BEVStereo \cite{li2023bevstereo} for extracting 3D features. For the gaussian head, we simply employed a two-layer fully connected network. For the RC module, we used only one future frame as the basic experimental configuration for all experiments. For balance weights, we follow the original RenderOcc configuration. For dynamic weights, we simply set $\alpha$ to 0.1 and $\omega$ to 0.8. During the training phase, we utilized the Adam optimizer with a linear learning rate scheduling strategy, set the batch size to 16, and trained for 12 epochs with a learning rate of 2e-4. All experiments were conducted on NVIDIA A6000 GPUs.

\subsection{Performance} We report the performance of GSRender in this part, demonstrating its superior results compared to the existing SOTA method. We achieved significant performance improvements in RayIoU. Building on these enhancements, we now turn to a detailed comparison, both qualitative and quantitative, to further validate the efficacy of our approach.

\subsubsection{Quantitative Comparison.} 
As illustrated in Table~\ref{tab:1}, The objective was to explore the performance of both methods when limited to a single-frame scenario, using mIoU as the evaluation metric. GSRender outperformed RenderOcc, achieving a higher mIoU, and enhanced performance when used as auxiliary supervision alongside 3D supervision, improving from \textbf{26.11 to 29.56}. When supervised only by the current frame, the inherent limitations of single view supervision cause both RenderOcc and GSRender to produce a thick surface. Consequently, relying solely on mIoU does not accurately reflect the model's performance. To truly assess the capabilities of our model, we adopted RayIoU \cite{tang2024sparseocc} as our primary evaluation metric to better evaluate the model's true capability in scene reconstruction. As shown in Table 2. it is surprising that by utilizing only one frame for auxiliary supervision. Our results achieved a significant improvement over RenderOcc which uses six auxiliary frames. In fact our approach even surpasses certain methods that rely on 3D occupancy ground truth such as SimpleOcc.

\subsubsection{Qualitative Comparison.} 
As illustrated in Figure~\ref{fig:5}, our experimental results demonstrate that the RC module significantly enhances the accuracy in positioning and shaping of specific categories, such as vehicles. In the third column, we visualized the results without using the RC module. This comparison clearly shows that reliance on single-frame ground truths tends to result in duplicated predictions. However, the integration of the RC module effectively addresses this problem, leading to a more unified and seamless reconstruction of the entire scene and enhancing the practicality of the prediction results.

\subsubsection{Ablations and Analysis.}
\label{sec:Analysis}
In our initial experiments, we postulated that using fixed Gaussian properties could also yield satisfactory rendering results. Nonetheless, this does not hold true. Therefore, we conducted ablation studies to verify the necessity of shifts in each Gaussian attribute, as shown in Table~\ref{tab:4},  indicate that Gaussian trained without mean and scale shift \( \delta_\mu, \delta_s\) and with rotation shift \(\delta_r\) exhibit decreased accuracy. Furthermore, in Table~\ref{tab:4}, we also validated the effectiveness of the dynamic weight \(\omega\) and the adjacent  weight \(\alpha\). Also presents the ablation experiment showing only the current frame as supervision, without the RC module. It can be seen that the RC module brings a significant improvement in performance. Notably, the mIoU for using only a single frame is comparable to that with the RC module, which is attributed to the hacked mIoU caused by duplicated predictions.

\begin{table}[t!]
\newcolumntype{a}{>{\columncolor{gray!7}}c}
\centering
\begin{tabularx}{0.85\linewidth}{c|ac}
\hline
                                  & RayIoU(\%) & mIoU(\%) \\ \hline
\multicolumn{1}{c|}{Ours}       &  \textbf{25.5}               &    \textbf{22.2}    \\
\multicolumn{1}{c|}{Ours w \(\delta_r\)}   &      24.0               &        22.1           \\
\multicolumn{1}{c|}{Ours w/o \(\delta_s\)}  &   8.5                &      6.2          \\
\multicolumn{1}{c|}{Ours w/o \(\delta_\mu\)} &  19.5                  &      16.1           \\ 
\multicolumn{1}{c|}{Ours \( |\delta_\mu| \leq 0.2\)} &  21.0                   &      18.7           \\  \hline
\end{tabularx}
\caption{Ablation of Each Gaussian Properties}
\vspace{-0.25cm}
\label{tab:3}  
\end{table}

\begin{table}[t!]
\newcolumntype{a}{>{\columncolor{gray!7}}c}
\centering
\begin{tabularx}{0.85\linewidth}{ccc|ac}
\hline
\(RC\) & \(\omega\) \quad &  \(\alpha\) \quad & RayIoU(\%)  & \quad mIoU(\%)  \\ \hline
$\checkmark$ & $\checkmark$ \quad & $\checkmark$ \quad & \textbf{25.5} & \textbf{22.2}  \\
$\checkmark$ & $\checkmark$ \quad & $\times $ \quad & 24.1 & 21.9 \\
$\checkmark$ & $\times$ \quad & $\checkmark$ \quad & 24.5 & 22.1 \\
$\checkmark$ & $\times$ \quad & $\times $ \quad & 23.5 & 21.8 \\
$\times$\quad  & $\times $ \quad & $\times$ \quad & 19.7 & 21.4 \\ \hline
\end{tabularx}
\caption{Ablation of RC module and loss weights.}
\vspace{-0.55cm}
\label{tab:4}
\end{table}

\begin{figure}[t!]
\includegraphics[width=\linewidth]{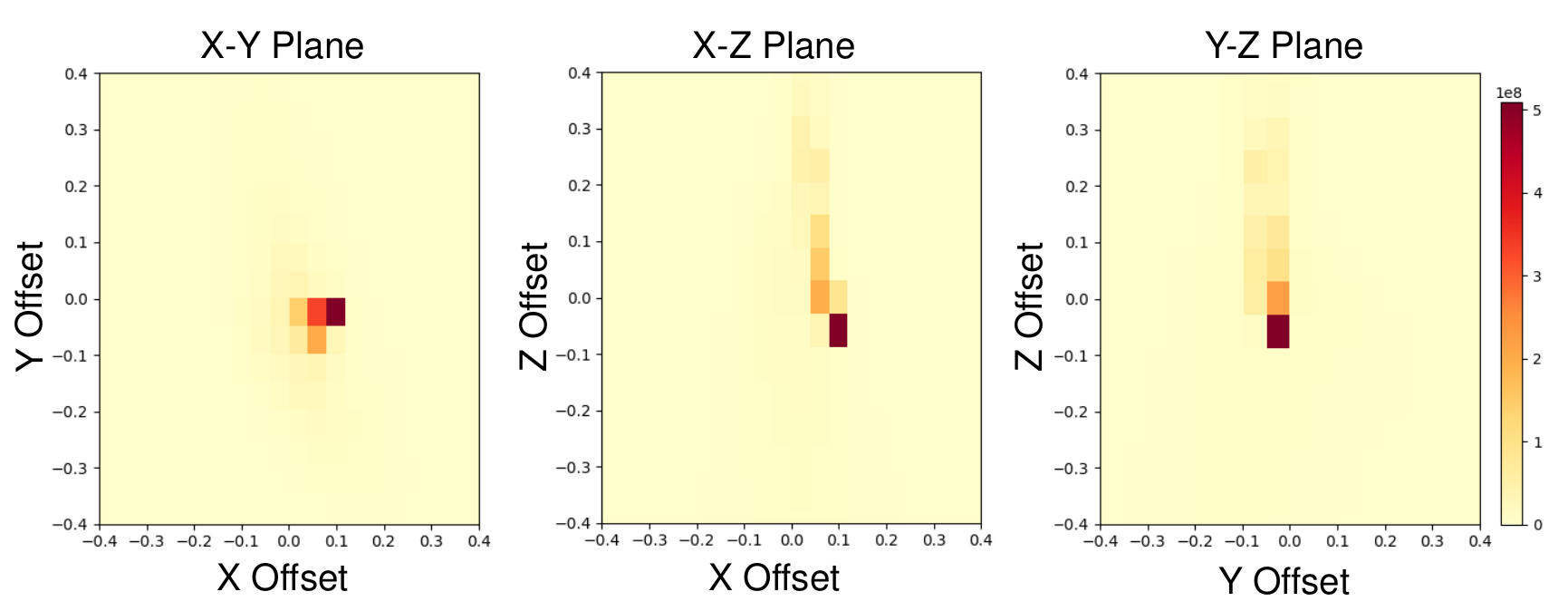}
    \caption{Statistics of  \(\delta_{\mu} \) distribution. Most Gaussians move within the voxel, with a notable deviation in the z-direction.}
    \label{fig:6}
\end{figure}

\noindent To better investigate the benefits of the Gaussian representation, we excluded Gaussians that do not correspond to any occupied space, then analyzed the distribution of mean shifts among the remaining Gaussians, as illustrated in Figure~\ref{fig:6}. Notably, the majority of these Gaussian means remained within their original positions, with shifts not exceeding half voxel. This stability enabled us during inference to directly associate the semantic labels of the Gaussians with their corresponding voxels, leading to robust results. 

\begin{figure}[t!]
\includegraphics[width=\linewidth]{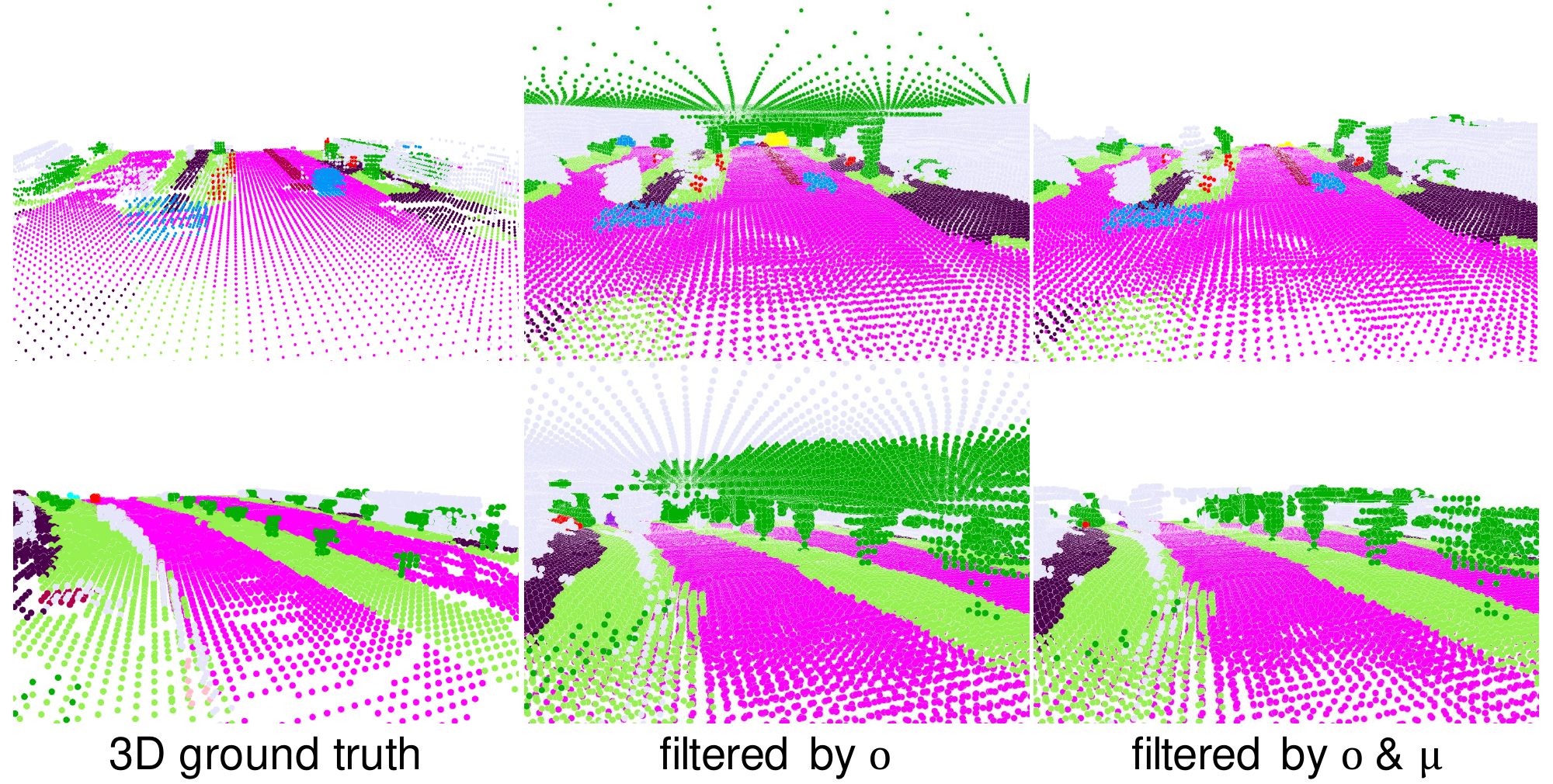}
    \caption{Comparison after filtering Gaussians.}
    \label{fig:7}  
\end{figure}

\noindent Furthermore, we visualized more intuitive outcomes, as demonstrated in the Figure~\ref{fig:7}, where the positions of Gaussian distributions were depicted as a point cloud.  The second column displays results filtered by opacity alone. Many Gaussians in the sky lack supervision, which results in Gaussians near object surfaces being categorized as adjacent objects and exhibiting a tendency to shift closer to these objects. This also confirms the common z-offset in many Gaussians exceeds half a voxel and shows, as in Table~\ref{tab:3}, that overly constraining the Gaussian mean \(\mu\) degrades performance. Building on this, as shown in the third column of Figure~\ref{fig:7}, we have additionally filtered out Gaussians that have shifted beyond the confines of their voxels. The results indicates that most Gaussians near real object surfaces adhered to their voxel positions. Additionally, since the 2D ground truth is directly derived from LiDAR point clouds, it more accurately reflects the distribution of objects in the real world compared to 3D ground truth. Consequently, our method achieves object shapes , like trees, that are more realistic than the 3D ground truth.


\section{Limitation and future work}
Qualitative results indicate that our method still has several limitations. We treat each voxel as a Gaussian, but in practice, such a large number of redundant Gaussians is unnecessary, leading to increased computational burden and memory usage. Furthermore, the sparsity of LiDAR supervision could also impede the optimization of Gaussians, thereby affecting the overall performance of the model. GSRender serves as a foundation for our future work, where we will focus on reducing the number of Gaussians to further optimize performance.  

\section{Conclusion}
In this work, we explored the application of 3D Gaussian Splatting in 3D occupancy prediction tasks. We implemented a method to mitigate the issue of duplicate predictions along the same camera ray when using 2D weakly supervision. Our experiments validated the effectiveness of our approach, achieving state-of-the-art performance among current 2D-supervised methods. And provided a detailed analysis for the performance improvement. This technique has the potential to become a powerful tool in both industry and academia.

\bibliographystyle{IEEEtran}
\bibliography{root} 

\setcounter{page}{1}

\section{Additional Experiments and Analysis}

\noindent \textbf{Sampling at the Logits Level. }
We also conducted additional experiments aimed at alleviating the computational burden of sampling at the feature level. Instead of extracting features for sampling, we opted to directly shift the Gaussian positions of the current frame. These shifted Gaussian spheres were then supervised using rendering results derived from auxiliary frames. This approach simplifies the sampling process and is conceptually equivalent to performing sampling directly on the logits, effectively bypassing any truncation errors that might arise in feature-level operations.
\begin{table}[ht]
\centering
\newcolumntype{a}{>{\columncolor{gray!7}}c}
\begin{tabularx}{0.88\columnwidth}{c|ac} 
\hline
& \quad RayIoU(\%) \quad & \quad mIoU(\%) \quad \\ \hline
\quad Feature \quad &\textbf{25.5} &  \textbf{22.2} \\
\quad Logits \quad &  19.7   &  21.3 \\ \hline
\end{tabularx}
\caption{Comparison of Different Frame Intervals. +1 represents the next immediate frame following the current frame.}
\label{tab:sup2}
\end{table}

However, the results revealed a key limitation: this direct and brute-force sampling method did not achieve the same level of performance as feature-level sampling, even though the latter inherently introduces truncation errors. This discrepancy suggests that feature-level sampling retains more nuanced information critical for accurate predictions. While direct sampling eliminates certain sources of error, it appears to lose the fine-grained contextual details embedded within the features, which are crucial for high-quality results.

Moreover, the brute-force nature of this approach may also limit its ability to adapt dynamically to variations in the data, making it less robust across diverse scenarios. These findings underscore the importance of feature-level sampling, where the inherent richness of the extracted features compensates for the challenges posed by truncation errors. This balance between computational simplicity and performance accuracy highlights the trade-offs involved in designing efficient and effective sampling methods for tasks requiring precise supervision.

\begin{table}[t!]
\newcolumntype{a}{>{\columncolor{gray!7}}c}
\centering
\begin{tabularx}{0.88\linewidth}{cccc|ac}
\hline
\(-2\) & \(-1\) &  \(+1\) & \(+2\) & RayIoU(\%)  & mIoU(\%)  \\ \hline
$\times$   & $\times$  & $\times$ & $\checkmark$ & 25.4 & 22.5 \\
$\checkmark$   & $\times$  & $\times$ & $\times$ & 25.2 & 21.4 \\
$\times$  & $\checkmark$ & $\times$ & $\times$ & 25.2 & 22.4 \\
$\times$  & $\times$  & $\checkmark$ & $\times$ & 25.5 & 22.2 \\
$\checkmark$  &  $\checkmark$ & $\times$ & $\times$ & 25.5 & 22.6 \\ 
$\times$ & $\times$  &  $\checkmark$  &  $\checkmark$  & 25.8 & 22.8 \\
$\checkmark$ & $\checkmark$  & $\checkmark$ & $\checkmark$ & 26.7 & 23.5 \\ \hline
\end{tabularx}
\caption{\textbf{Comparison of Different Frame Intervals.} +1 represents the next immediate frame following the current frame}
\label{tab:sup1}
\end{table}
\begin{figure*}[t!]
    \includegraphics[width=\linewidth]{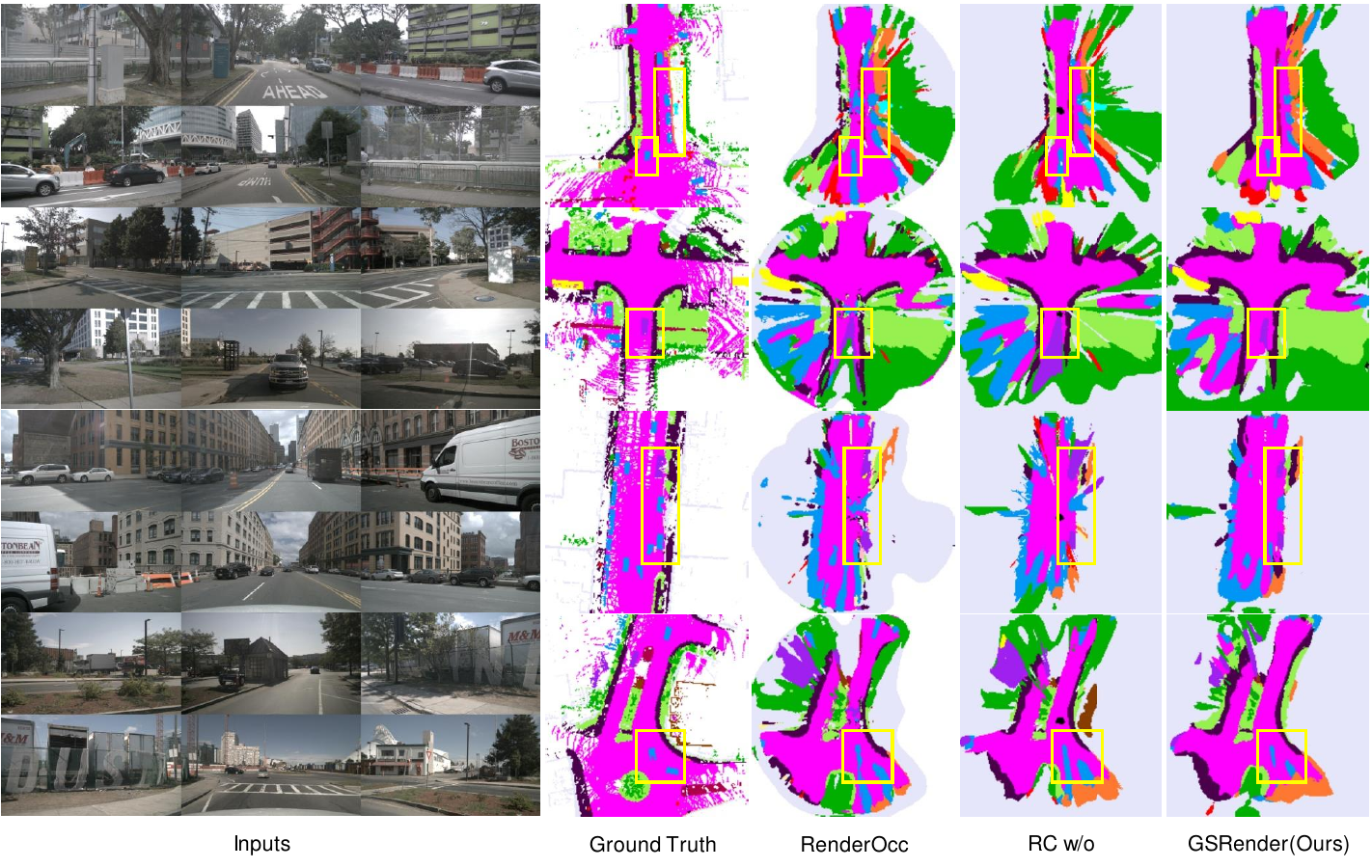}
    \caption{\textbf{Additional qualitative results.}. Top view on Occ3D-nuScenes}
    \label{fig:supexp}
\end{figure*}
\noindent \textbf{Comparison of Different Frame Intervals.} In the experimental section of the main text, we used only a single frame of future auxiliary supervision to simplify the experimental setup and focus on the core performance metrics. To gain a deeper understanding of how different auxiliary frame intervals impact the results, we conducted a series of additional experiments, as summarized in Table \ref{tab:sup1}. These experiments systematically evaluated the performance using various combinations of auxiliary frames, including the next frame, the previous frame, the second previous frame, the second future frame, the previous two frames, the future two frames, and a combination of both previous and future two frames.

The results of these experiments highlighted several notable trends. First, using only a single auxiliary frame, whether from the past or the future, produced minimal improvements over the baseline, suggesting that the impact of a single frame is relatively limited. When two auxiliary frames were combined, there was a slight but noticeable increase in accuracy, although the improvement was still not particularly significant. This indicates that while incorporating additional temporal context can enhance performance, the benefits are relatively modest on this scale.

To provide a fair and thorough comparison with RenderOcc, which operates under a more comprehensive experimental setup, we chose to use an extended configuration in the main text. However, in practical scenarios, increasing the number of accumulated auxiliary frames typically results in a more substantial boost in accuracy. This improvement becomes particularly evident as more frames are included, as they provide richer temporal context and more robust supervision for tasks such as occupancy prediction. However, the rate of improvement diminishes with each additional frame, suggesting diminishing returns as temporal information becomes increasingly redundant.

It is also important to consider the computational trade-offs. While using multiple auxiliary frames enhances accuracy, it comes at the cost of increased computational complexity and reduced efficiency. The additional frames require more processing power and memory, potentially limiting real-time applications where efficiency is critical. This trade-off is one of the key reasons why we opted to present results based on a single auxiliary frame in the main text, striking a balance between accuracy and computational feasibility.

\begin{figure}[t!]
\includegraphics[width=\linewidth]{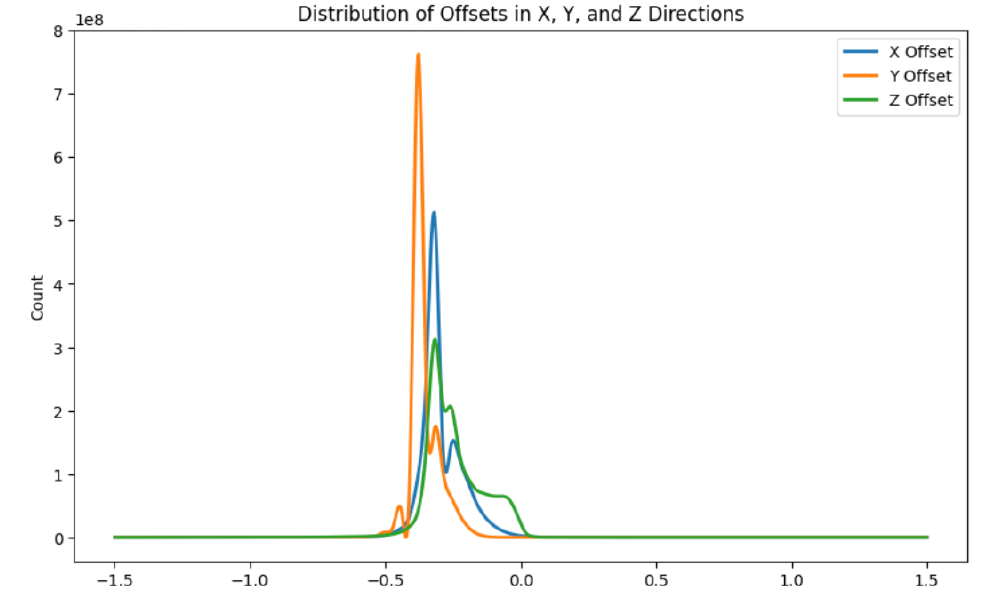}
    \caption{Statistics on the scales \(s\).}
    \label{fig:sup1}  
\end{figure}

\noindent \textbf{Statistics on the scales s.} As illustrated in the figure \ref{fig:sup1}. For all voxels classified as occupied by the network, a detailed analysis of the scale revealed that the scales of most filtered Gaussians exhibited minimal variation, with the majority maintaining a consistent size. This observation suggests that the voxel resolution is too coarse to adequately represent fine-grained details, making it difficult for the Gaussians to capture accurate shapes. In other words, the current voxel size likely smooths over intricate structural variations, leading to a loss of precision in modeling.

Moreover, the weakly supervised labels derived from LiDAR projections introduce additional challenges. These labels, which are mapped from sparse 3D points to 2D pixel space, often exhibit discontinuities in semantic consistency across neighboring pixels. Such irregularities in the labeling process can further hinder the Gaussians from conforming to their true shapes, as they rely on coherent supervision for accurate fitting. Consequently, some Gaussians may fail to achieve the desired opacity distribution, either overestimating empty regions as occupied or misrepresenting the shape of actual objects.

\noindent \textbf{Additional qualitative results.} Due to space limitations in the main text, we present additional experimental results here to further validate the effectiveness of our method. As shown in \ref{fig:5}.

\section{Specific experimental configuration.}
We did not specifically fine-tune the weights for dynamic objects or neighboring frames. Instead, we adopted a straightforward approach by reducing the depth contribution of dynamic objects, setting the weight parameter \(\alpha\) to 0.1. To ensure the supervision from the current frame remains dominant, we assigned a weight \(\omega\) of 0.8 to neighboring frames, prioritizing the current frame’s information during training.

In addition to this weighting scheme, we applied constraints to the Gaussian center offsets, limiting them to a range of three adjacent voxels. This decision was informed by observations detailed in the main text, which highlight the trade-offs in offset range selection. Overly restricting the offset range can hinder the ability of Gaussians to achieve accurate opacity, leading to misclassifications where empty spaces are mistakenly treated as occupied voxels. Conversely, completely removing restrictions on the offset range can result in gradient explosions, destabilizing the training process.

To balance these extremes, we adopted a relaxed constraint, permitting Gaussian offsets to operate within a three-voxel radius in each direction. This approach strikes a balance by providing sufficient flexibility for the Gaussians to adapt to local variations while avoiding instability. 

\section{Additional Limitations and Discussion}
As previously discussed and confirmed by experimental results, the use of 3D Gaussian splatting effectively mitigates the sensitivity of accuracy to the number of samples. However, we still aim to achieve a more sparse 3D Gaussian representation based on the LSS method. Specifically, we envision representing the entire 3D space objects with only sparse 3D Gaussians, ultimately deriving a more refined occupancy representation through the Gaussian-to-Voxel module. This remains a key objective for our future work.

During our exploration, we observed that downstream tasks cannot accommodate intermediate states with duplicate predictions. To address this, we introduced the Ray Compensation module, which significantly alleviates the issue of duplicate predictions. This module is not limited to networks based on 3D Gaussian Splatting (3DGS). We believe it can serve as a general-purpose module that can be integrated into any model. By simply sampling the features of adjacent frames and using the labels of those frames for supervision, it can achieve promising results. This functionality constitutes a core innovation of our work, with potential for adoption in both industrial and academic applications.

\section{A Broader impacts}
Our research delves into a practical, end-to-end weakly supervised training framework based on 3D Gaussian Splatting(GS). We have introduced new modules that help reduce the redundancy in 3D space predictions caused by supervising with 2D projections of 3D LiDAR data. This approach not only boosts the performance of 2D supervision alone, but also, when used as auxiliary 3D supervision, shows even better results. It is a fresh solution for both the industry and academia, and it is vital for autonomous driving systems. By integrating predictions from auxiliary supervision into downstream tasks, we can provide more robust conditions for these tasks.

In the real world of autonomous driving, the accuracy of predictions and the system's ability to respond in real-time are both critical. Incorrect predictions or slow response times can lead to safety risks. Since autonomous driving systems are directly linked to personal safety, improving the network's real-time capabilities and developing more precise prediction methods are essential. The current bottleneck in network latency is the backbone network itself. Therefore, our next steps are to streamline the backbone network by reducing redundancy and to design networks that are more responsive.

\section{Licenses for involved assets}
Our code is built on top of the codebase provided by RenderOcc\cite{pan2024renderocc} and BEVStereo\cite{li2023bevstereo}, which is subject to the MIT license.
Our experiments are carried out on the Occ3D-nuScenes\cite{tian2024occ3d} which provides occupancy labels for the nuScenes dataset\cite{caesar2020nuscenes}. Occ3D-nuScenes is licensed under the MIT license and nuScenes is licensed under the CC BY-NC-SA 4.0 license.

\end{document}